\documentclass[10pt,twocolumn,letterpaper]{article}

\usepackage{cvpr}              % To produce the CAMERA-READY version
\usepackage{multirow}

\usepackage[dvipsnames]{xcolor}

\definecolor{cvprblue}{rgb}{0.21,0.49,0.74}
\usepackage[pagebackref,breaklinks,colorlinks,citecolor=green]{hyperref}

\def\paperID{*****} % *** Enter the Paper ID here
\def\confName{CVPR}
\def\confYear{2024}

\title{TextCoT: Zoom In for Enhanced Multimodal Text-Rich Image Understanding}

\author{Bozhi~Luan$^{1,}$\thanks{Equal contribution.}~~~~~Hao~Feng$^{1,*}$~~~~~Hong~Chen$^{2}$~~~~~Yonghui~Wang$^{1}$~~~~~Wengang~Zhou$^{1,}$\thanks{Corresponding authors: Wengang Zhou and Houqiang Li.}~~~~Houqiang~Li$^{1,}$\footnotemark[2] \\
 	{\normalsize $^{1}$ University of Science and Technology of China}, 
  	{\normalsize $^{2}$ Merchants Union Consumer Finance Company Limited} \\
	{\tt\small \{bzluan,haof,wyh1998\}@mail.ustc.edu.cn, \{zhwg,lihq\}@ustc.edu.cn,} \
\tt\small chenhong@mucfc.com}

\begin{document}
\maketitle

\begin{abstract}
The advent of Large Multimodal Models (LMMs) has sparked a surge in research aimed at harnessing their remarkable reasoning abilities. However, for understanding text-rich images, challenges persist in fully leveraging the potential of LMMs, and existing methods struggle with effectively processing high-resolution images. In this work, we propose TextCoT, a novel Chain-of-Thought framework for text-rich image understanding. 
TextCoT utilizes the captioning ability of LMMs to grasp the global context of the image and the grounding capability to examine local textual regions.
This allows for the extraction of both global and local visual information, facilitating more accurate question-answering.
Technically, TextCoT consists of three stages, including image overview, coarse localization, and fine-grained observation.
The image overview stage provides a comprehensive understanding of the global scene information, and the coarse localization stage approximates the image area containing the answer based on the question asked.
Then, integrating the obtained global image descriptions, the final stage further examines specific regions to provide accurate answers.
Our method is free of extra training, offering immediate plug-and-play functionality.
Extensive experiments are conducted on a series of text-rich image question-answering benchmarks on several advanced LMMs, and the results demonstrate the effectiveness and strong generalization ability of our method. Code is available at \url{https://github.com/bzluan/TextCoT}.
\end{abstract}

\section{Introduction}

\begin{figure}[t]
  \centering
  \includegraphics[width=\linewidth]{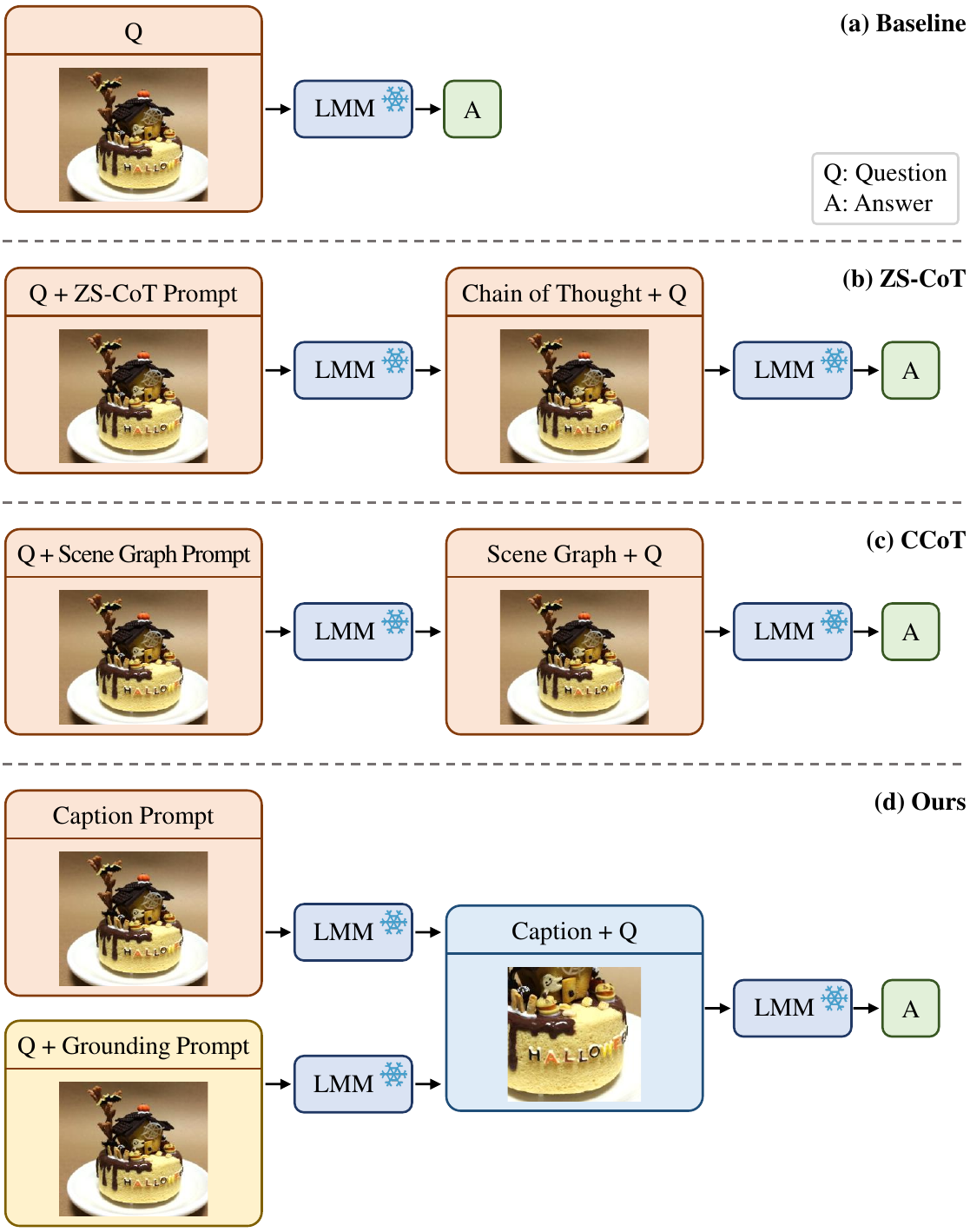}
  \caption{
     A pipeline comparison of (a) baseline LMM, (b) Zero-shot CoT~\cite{zs-CoT}, (c) CCoT~\cite{cCoT}, and (d) our proposed TextCoT. For enhanced comprehension of text-rich images, TextCoT leverages the captioning and grounding abilities of LMMs to grasp the global context and local textual regions 
     within the image, respectively.
  }
  \label{fig:figure1}
\end{figure}

In recent years, breakthroughs in the field of Large Language Models (LLMs) have revolutionized the Natural Language Processing (NLP) field with impressive performance across a wide range of tasks.
Building upon the foundational LLMs, Large Multimodal Models (LMMs) are developed by integrating visual information into LLMs, enabling them to acquire information from both images and text. Research on LMMs continually improves model architecture, training data, and training strategies, etc., thereby enhancing performance in various scenes.
In the domain of multimodal understanding, research on text-rich images represents a critical direction. Distinguished from general image understanding, the challenge of this task lies in the necessity for models to concurrently comprehend the intertwined visual and textual contents within images. Additionally, due to the presence of text, such images often feature higher resolutions, as exemplified by document images \cite{document}.

In the field of text-rich image understanding, general large multimodal models~\cite{ye2023mplugowl,llava,llava1.5} often exhibit suboptimal performance. This can be attributed to the fact that the answers to questions posed by such tasks are typically textual and are frequently located in local regions of the image, such as identifying the license plate number of a car within a high-resolution image. However, the input resolution of existing LMMs is usually constrained to be below 448 pixels, which inhibits their ability to effectively address these types of queries.
To tackle this problem, some approaches opt to develop high-resolution visual encoder~\cite{feng2023docpedia} or to divide images into multiple patches~\cite{li2023monkey,ye2023ureader} for individual encoding of visual features. Nevertheless, such solutions necessitate additional training and demand substantial resources for the collection and construction of high-resolution image-text question-answering datasets.

Within the domain of large language models, research based on the Chain-of-Thought (CoT) techniques aims to overcome the problem by deliberate reasoning steps without extra model training, and they have achieved amazing performance enhancement. 
These techniques can be directly applied to our text-rich image understanding field. 
As illustrated in Figure~\ref{fig:figure1}(b), the representative methodology ZS-CoT~\cite{zs-CoT} incorporates the process of step-by-step reasoning as a new input to the model, thereby obtaining more accurate responses. However, ZS-CoT~\cite{zs-CoT} does not substantively address the issue that this task necessitates a meticulous examination of localized regions.
Furthermore,
CCoT~\cite{cCoT} represents an advanced CoT method tailored for LMMs. It enhances question-answering capabilities in general scenes by constructing a scene graph that models the relationships among objects, as shown in Figure~\ref{fig:figure1}(c). However, CCoT~\cite{cCoT} is noted to perform suboptimally on text-rich images. This can be attributed to the challenges associated with constructing scene graphs for textual instances.

To address the above issues, in this work we propose TextCoT, a novel Chain-of-Thought framework for text-rich image understanding. 
Our idea is inspired by the human cognitive patterns. Concretely, in a text-rich scenario where it is unfeasible to remember all details, people rarely actively predict which part to remember for future questioning.   
Instead, a more intuitive strategy is scanning the text guided by the question to pinpoint areas likely to contain the answer, followed by a closer examination of details to formulate a response.
To emulate this multimodal thought process, we developed our TextCoT.
Technically, our TextCoT consists of three stages, including image overview, coarse localization, and fine-grained observation.
The initial stage of image overview facilitates a thorough grasp of the overarching scene information, while the subsequent coarse localization stage estimates the section of the image that encompasses the answer, contingent upon the posed question. Following this, by assimilating the global descriptions of the image garnered earlier, the concluding phase delves into particular areas in greater detail to furnish precise answers. 

To verify our TextCoT, we perform extensive experiments on a series of text-rich image question-answering benchmark datasets based on several advanced LMMs. The quantitative and qualitative
results exhibit the effectiveness and strong generalization ability of our method. 
We summarize our contributions as follows:
\begin{itemize}
    \item We propose TextCoT, a novel Chain-of-Thought framework for text-rich image understanding. TextCoT employs the grounding capabilities of LMMs to examine specific answer regions, thereby facilitating more accurate and nuanced question-answering.
    \item We attempt to address the issue of high-resolution visual inputs within the task of text-rich image understanding from a perspective of Chain-of-Thought. Our method is free of extra training and is ready for plug-and-play.
    \item We perform extensive experiments on a series of text-rich image question-answering benchmark datasets based on several advanced LMMs to verify our methods.
\end{itemize}

\section{Related Work} 
Our TextCoT is a multimodal Chain-of-Thought approach, aiming to enhance the performance of Large Multimodal Models (LMMs) in text-rich scenarios by effectively utilizing the capabilities of LMMs. 
In the following, we first review the research on Large Language Models (LLMs) and LMMs. Then we discuss the literature about Chain-of-Thought (CoT) for LLMs and LMMs.

\smallskip\textbf{Large Language Models (LLMs)}. 
With the demonstration of the powerful potential of the Transformer architecture, Large Language Models (LLMs) radically transformed the field of Natural Language Processing (NLP). Models like BERT~\cite{devlin2018bert} and T5~\cite{t5}, employing encoder-decoder architectures, laid the foundation for a deep understanding of linguistic nuances. GPT3~\cite{gpt3}, with its decoder-centric design, showcased exceptional performance in few-shot and zero-shot learning scenarios, illustrating adaptability across a wide range of NLP tasks. The PaLM~\cite{palm} model, by expanding model parameters and dataset breadth, pushed the limits of comprehension and generative capabilities. InstructGPT~\cite{instructgpt} and ChatGPT~\cite{chatgpt} introduced fine-tuning and reinforcement learning from human feedback, significantly enhancing the quality of interactions. Following these, open-source models such as LLaMA~\cite{llama} and Vicuna~\cite{chiang2023vicuna} continued to advance the frontier benchmarks of NLP, paving new pathways for future research.

\begin{figure*}[t]
  \centering
  \includegraphics[width=\linewidth]{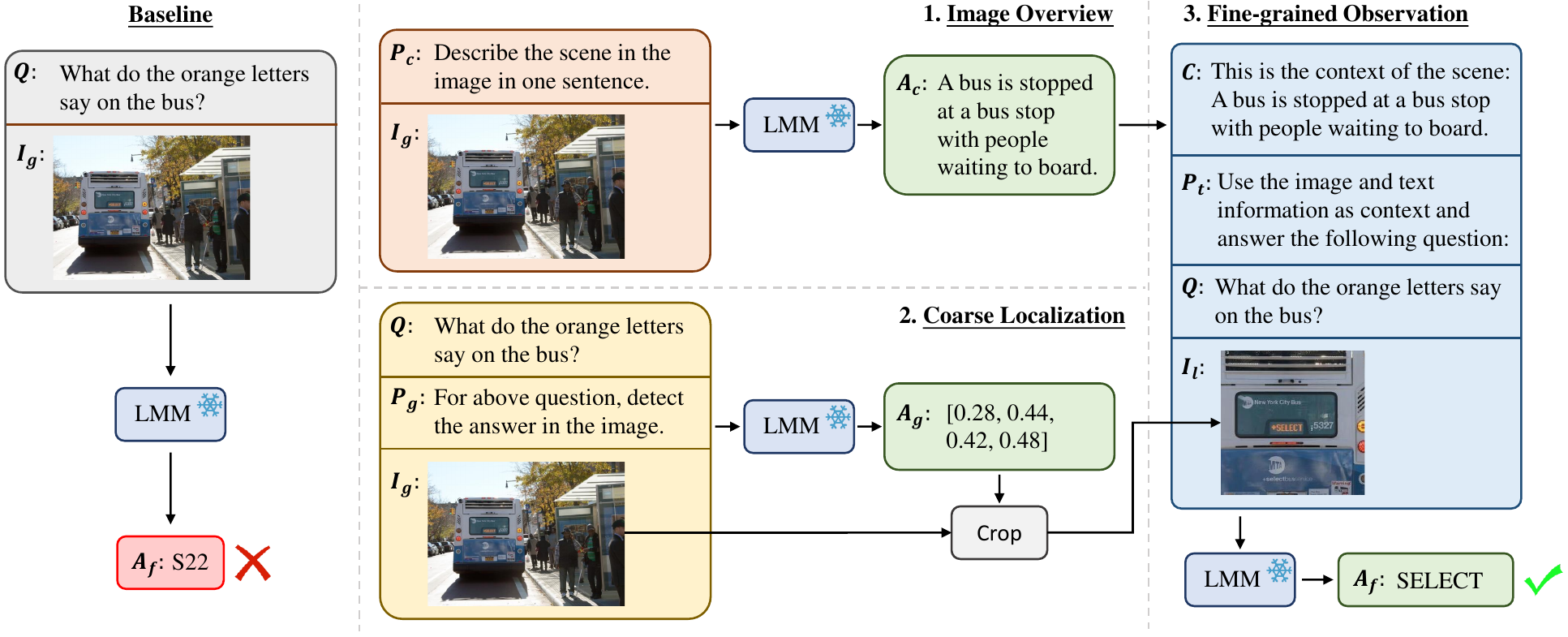}
  
  \caption{An overview of the standard one-stage LMM (left) and our TextCoT (right).  
  TextCoT comprises three stages: (1) Image Overview, (2) Coarse Localization, and (3) Fine-grained Observation. The initial two phases respectively generate a global context description \(A_c\) of image \(I_g\) and an answer region \(A_g\) to question \(Q\), 
  facilitating the production of a more accurate response \(A_f\).
  }
  \label{fig:framework}
\end{figure*}

\smallskip\textbf{Large Multimodal Models (LMMs).} 
To incorporate visual knowledge into Large Language Models (LLMs), large multimodal models (LMMs) were developed by integrating the visual encoders of existing vision-language models \cite{clip, alphaclip, li2022blip} with the reasoning capabilities of LLMs. Through improved pre-training alignment and instruction fine-tuning, numerous studies such as MiniGPT-4 \cite{minigpt4}, LLaVA \cite{llava,llava1.5}, InstructBLIP \cite{instructblip}, and mPLUG-Owl \cite{ye2023mplugowl} demonstrated significant progress in various vision-language tasks. 
Some studies \cite{chen2023shikra,peng2023kosmos,huang2023opera,textgrounding} leveraged object detection and text detection data to enhance the grounding capabilities of LMMs and reduce hallucinations. 
ShareGPT4V \cite{sharegpt4v}  improved the alignment between image and text modalities by enhancing the quality of model caption data. Vary \cite{wei2023vary}, V* \cite{guidedvisualsearch}, and DualFocus\cite{cao2024dualfocus} further 
raised the benchmarks for LMMs through improvements of the model architecture and training framework.

Many LMMs perform poorly in text-rich scenarios due to the dense fine-grained information and the high image resolution. To address this issue, studies like UniDoc \cite{feng2023unidoc} and mPLUG-DocOwl \cite{mplugdocowl} utilized text-related visual task datasets to enhance model capabilities in text-rich scenarios. Vary-toy \cite{varytoy} and Qwen-VL \cite{bai2023qwenvl} enhanced the ability to high resolution image understanding by training larger visual encoders. UReader \cite{ye2023ureader}, Monkey \cite{li2023monkey} \cite{liu2024textmonkey}  adopted the method of stacking multiple vision encoders to increase model input resolution. DocPedia \cite{feng2023docpedia} proposed using frequency domain visual information to expand input resolution while reducing token usage. These methods significantly improved the models' input resolutions, greatly enhancing their ability to understand fine-grained details such as text. Although these approaches showed excellent results through training for more complex model architectures and visual tasks, the dependence on high-quality training data for visual instruction remains a significant challenge.

\smallskip\textbf{Chain-of-Thought Prompting.}
A series of studies focused on the Chain-of-Thought (CoT) Prompting methods highlighted the significant potential of large language models, while also uncovering the fact that their performance is not fully realized due to inadequate prompt techniques. 
CoT methods control the LLMs and LMMs through prompts during the inference stage, eliciting the reasoning potential of the model without the need for training or fine-tuning.
Research endeavors such as CoT~\cite{CoT}, Zero-shot CoT \cite{zs-CoT}, CoT-SC \cite{CoT-sc}, TOT \cite{tot}, and GOT \cite{got} revealed substantial enhancements in the reasoning capabilities of LLMs and laid the groundwork for Chain-of-Thought Prompting.

Many research \cite{acmmmprompttuning, acmmmreasoningprompt, acmmmrelationprompt} are dedicated to the precise manipulation of prompts and training processes to enhance the reasoning ability of vision language models. 
With the emergence of LMMs, many studies focusing on the CoT method of LMMs have been developed to enhance their reasoning ability. 
VidIL \cite{vidil}, DDCoT \cite{zheng2023ddCoT} and Multimodal-CoT \cite{mmCoT} are developed utilizing the captioning and reasoning abilities of LMMs and achieved promising results.
CCoT~\cite{cCoT} proposed utilizing JSON format to generate scene graphs, greatly enhancing the understanding of LMMs about the object relationships in the images.
CoCoT \cite{zhang2024coCoT} enhances the reasoning abilities of LMMs by observing contrastive information between multiple images.
Some studies \cite{wu2024dettoolchain, lei2024scaffolding} enhance the detection ability of LMMs by adding grids and dot matrices on the image. 

\begin{figure*}[t]
  \centering
  \includegraphics[width=\linewidth]{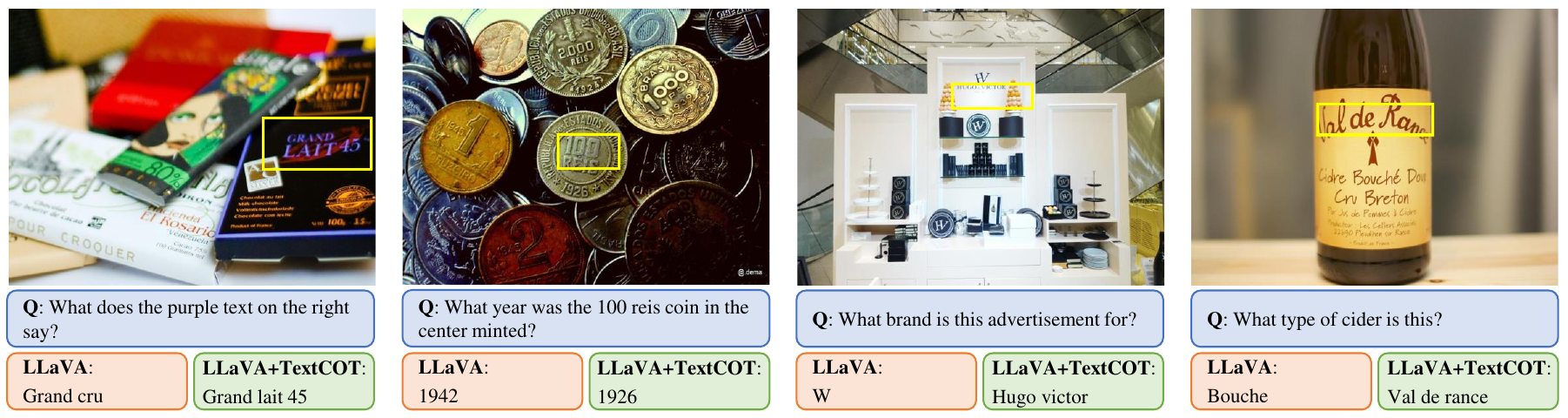}
  
  \caption{
    Comparison between the responses of LLaVA~\cite{llava1.5} and those augmented by our TextCoT on Scene Text-Centric VQA datasets. The estimated answer regions \(A_g\) in the second stage are highlighted in the image using yellow bounding boxes.
  }
  \label{fig:scenetext_case_1}
\end{figure*}

\begin{figure*}[t]
  \centering
  \includegraphics[width=\linewidth]{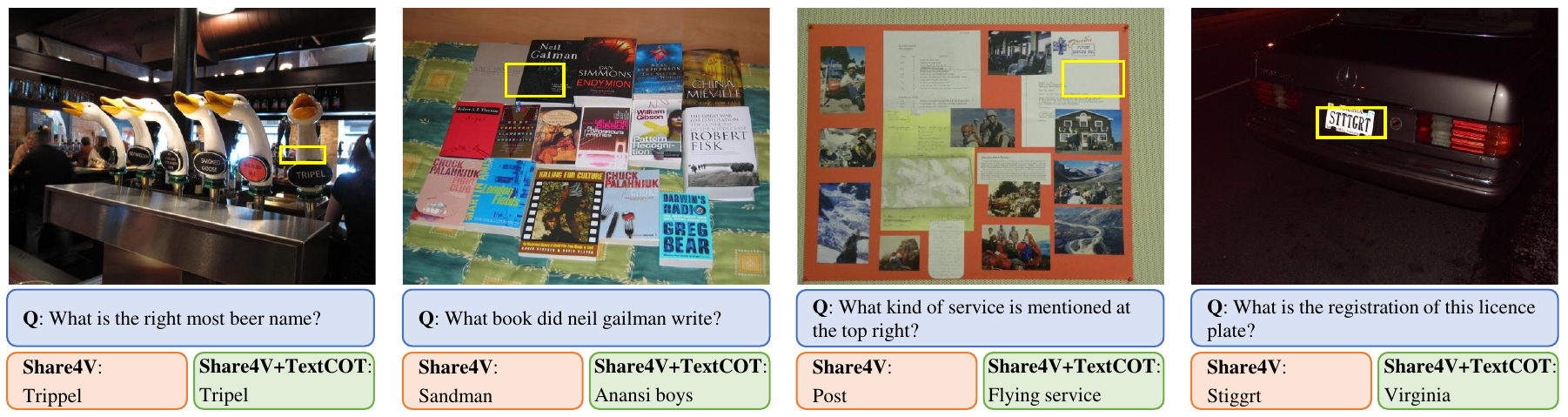}
  
  \caption{
      Comparison between the responses of ShareGPT4V~\cite{sharegpt4v} and those augmented by our TextCoT on Scene Text-Centric VQA datasets. The estimated answer regions \(A_g\) in the second stage are highlighted in the image using yellow bounding boxes.
  }
  \label{fig:scenetext_case_2}
\end{figure*}

Although these works excel in various aspects, they each have their flaws. Regular LMMs suffer from low resolution, leading to poor performance in textual scenarios due to their inability to capture fine details. The design, training, and fine-tuning of high-resolution models require significant resources and often lead to performance decreases on long-text conversation and localization tasks.  
Current multimodal CoT methods fail to address the critical issues in text-rich scenarios and the exploration of cross-modality prompting is insufficient.
This highlights a current research gap: the development of a multimodal CoT method capable of reasoning across both visual and textual modalities.

\section{Method}
As shown in Figure~\ref{fig:framework} (left), given a global text-rich image \(I_g\) and a question \(Q\), a straightforward approach is to directly instruct the LMMs to generate an answer \(A_f\). However, constrained by the granularity of the visual input, the LMMs often struggle with providing accurate responses. In this work, we propose to employ LMMs to examine specific regions, thereby facilitating more accurate question-answering. 
We present TextCoT, a novel Chain-of-Thought framework for text-rich image understanding.
Figure \ref{fig:framework} (right) presents an overview of TextCoT. In the following, we introduce its three stages, including (1) Image Overview, (2) Coarse Localization, and (3) Fine-grained Observation.

\subsection{Image Overview}

Our first step aims to leverage the captioning ability of LMMs to generate a concise yet comprehensive description of the image, thereby retaining the overall information within the image. 
Specifically, as shown in Figure~\ref{fig:framework}, 
we employ the global image \(I_g\) and the captioning prompt \(P_c\) to instruct the LMM and obtain a caption answer \(A_c\). This \(A_c\) will later be transformed into a caption \(C\) in the fine-grained observation stage, which provides global context to assist the answering of the given question.

Recent studies~\cite{yin2023woodpecker,wang2024vigc} have conclusively shown that longer outputs from Large Multimodal Models tend to exhibit increased hallucination, while longer input prompts also degrade performance. Consequently, as shown in Figure~\ref{fig:framework}, we incorporate the phrase  ``in one sentence'' into the prompt \(P_c\) to regulate the caption length. In this way, we encourage the generated caption to be concise, accurate, and able to summarize the scenario depicted in the image.

\begin{figure*}[h]
  \centering
  \includegraphics[width=\linewidth]{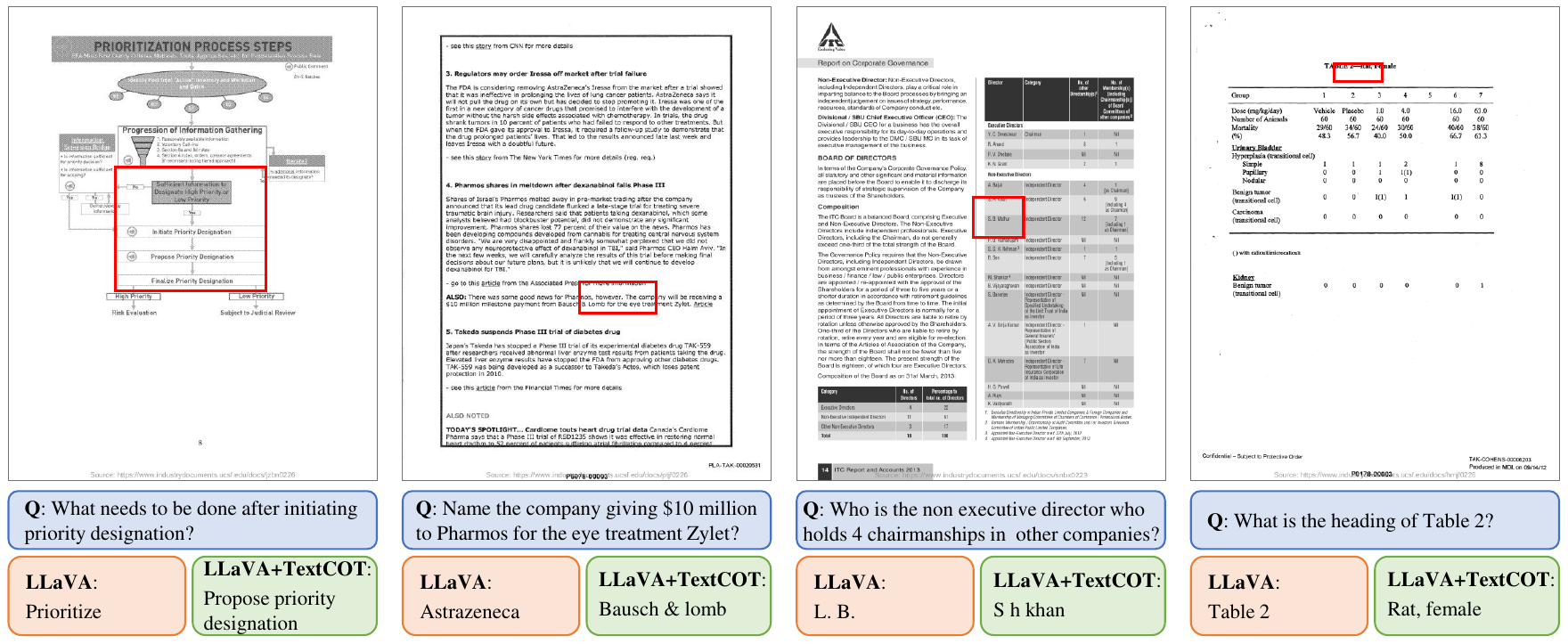}
  
  \caption{
      Comparison between the responses of LLaVA~\cite{llava1.5} and those augmented by our TextCoT on Document-Oriented VQA datasets. The estimated answer regions \(A_g\) in the second stage are highlighted in the image using red bounding boxes.
  }
  \label{fig:document_case}
\end{figure*}

\subsection{Coarse Localization}

Our second step aims to localize the answer in image \(I_g\) utilizing the grounding ability of LMMs. 
Concretely, as shown in Figure \ref{fig:framework}, we employ the question \(Q\), the grounding prompt \(P_g\), and the global image \(I_g\) to instruct the LMM, which generates a grounding answer \(A_g\). This grounding answer includes the bounding box coordinates of the answer. Subsequently, we crop the image based on the bounding box to obtain a local image \(I_l\). \(I_l\) is utilized in the third phase to enable the LMM to meticulously examine this specified area, thereby generating the correct answer to the question \(Q\). We introduce our cropping strategy next.

As shown in Figure \ref{fig:framework}, 
given a global image \(I_g\) and a grounding answer \(A_g\), we extend its shorter sides to match the longer side, using the center of bounding box as the focal point. 
Thus we obtain a square bounding box and avoid the distortion caused by the resize operation of CLIP-ViT \cite{clip, exploringclip}. 
Then, we introduce a hyperparameter: expand ratio $\alpha$.
It is used to scale the cropped square and empirically set to 1.5.
Meanwhile, as the typical input resolution for LMMs is either 336\(\times\)336 or 448\(\times\)448, we establish a minimum image dimension of 448$\times$448, to prevent very small bounding boxes after the expanding operation.
After the aforementioned procedure, we obtain a localized region image \(I_l\) that contains the answer, ensuring it encompasses sufficient information to provide the correct response in the subsequent stage. 
If the square crop exceeds image boundaries, we shift it to keep within the image.

Many existing LMMs were not trained using datasets specifically designed for text detection. Their grounding capability is typically learned through training on object detection datasets like RefCOCO \cite{refcoco_1, refcoco_2}. As a result, inaccuracies in localization often occur when LMMs are prompted to locate text in an image. However, this limitation does not hinder our method. Our TextCoT does not demand exact location outputs and an approximate location is adequate.

\subsection{Fine-grained Observation}

The final stage examines the obtained answer region \(I_l\) to generate an accurate response to the question. 
Specifically, as described in Figure \ref{fig:framework}, 
we first prepend a prompt ``This is the context of the scene:'' before captioning answer \(A_c\) to form the caption \(C\). The question \(Q\) remains the same as in the second stage. The task prompt \(P_t\) serves to further prompt the model to pay attention to the above context \(C\) and the question \(Q\). Our final text prompt is composed of \(C\), \(P_t\), and \(Q\). Then, we instruct the LMM with this text prompt and the image \(I_l\). Finally, the LMM delivers the final accurate answer by integrating the global contextual information from image \(I_G\) with the fine-grained visual details from the local image \(I_l\), in response to the posed question \(Q\).

\begin{table*}[h]
\setlength{\tabcolsep}{1.65mm}
\centering
\small
  \caption{Qualitative comparison between the five advanced LMMs and those augmented by TextCoT on various VQA benchmarks. 
  }
  \label{tab:main_table}
  
  \begin{tabular}{l|cc|ccc|ccc|c}
    \toprule
    \multirow{2}{*}{Method} & \multicolumn{2}{c}{Scene Text-Centric VQA} &  \multicolumn{3}{c}{Document-Oriented VQA} & \multicolumn{3}{c|}{KIE} &  \multirow{2}{*}{Average}  \\
     & TextVQA & STVQA  & DocVQA & InfoVQA & ChartQA & POIE & SROIE & FUNSD\\
    
    \midrule
    LLaVA-1.5-7B \cite{llava1.5} & 37.92 & 38.85 & 8.97 & 14.71 & 6.40 & \textbf{2.61} & 1.73 & 0.46 & 13.96\\
    LLaVA-1.5-7B + \textbf{TextCoT} &  \textbf{46.08} & \textbf{40.69} & \textbf{11.35} & \textbf{15.43} & \textbf{6.74} & 2.20 & \textbf{6.20} & \textbf{3.02} & \textbf{16.46}\\

    \midrule
    LLaVA-1.5-13B \cite{llava1.5} & 41.14   & 41.96 & 10.17  & 17.17 & \textbf{7.19} & 3.21 & 2.31  & 0.93 & 15.51\\
    LLaVA-1.5-13B + \textbf{TextCoT} & \textbf{49.63} & \textbf{43.61} & \textbf{15.94} & \textbf{18.66}  & 6.74 & \textbf{3.24} & \textbf{5.33} & \textbf{10.67} & \textbf{19.23}\\

    \midrule
    SPHINX \cite{lin2023sphinx}& 36.42 & 36.96 & 6.51 & 15.13 & 5.73 & 3.68 & 1.37 & 0.23 & 13.25\\
    SPHINX + \textbf{TextCoT} & \textbf{47.18} & \textbf{39.04} & \textbf{8.47} & \textbf{15.91} & \textbf{5.96} & \textbf{4.28} & \textbf{2.59} & \textbf{2.09} & \textbf{15.69}\\
    
    \midrule
    ShareGPT4V \cite{sharegpt4v} & 43.10 & 43.75 & 11.56 & 15.19 & 8.65 & 10.97 & 6.84 & 1.86 & 17.74\\
    ShareGPT4V + \textbf{TextCoT} & \textbf{52.79} & \textbf{46.68} & \textbf{16.89} & \textbf{16.99} & \textbf{12.13} & \textbf{12.54} & \textbf{13.69} & \textbf{13.92} & \textbf{23.20}\\

    \midrule
    Qwen-VL-Chat \cite{bai2023qwenvl} & 62.17 & 53.65 & \textbf{48.24} & 23.44 & \textbf{72.02} & 17.54 & 34.22 & 20.65 & 41.49\\
    Qwen-VL-Chat + \textbf{TextCoT} & \textbf{64.32} & \textbf{55.02} & 45.97 & \textbf{25.72} & 68.76 & \textbf{19.21} & \textbf{40.20} & \textbf{27.61} & \textbf{43.35}\\
    
    \bottomrule
  \end{tabular}
\end{table*}

\section{Experiments}

In this section, we conduct extensive experiments on a series of text-rich image question-answering benchmark datasets based on several advanced LMMs. In the following, we first introduce these involved LMMs and benchmark datasets.
Furthermore, we present and discuss the experiment results and ablation study.

\subsection{Baseline LMMs}
In our experiments, we evaluate our TextCoT based on five well-known LMMs, including LLaVA-1.5-7B~\cite{llava1.5}, LLaVA-1.5-13B \cite{llava1.5}, SPHINX~\cite{lin2023sphinx}, ShareGPT4V \cite{sharegpt4v}, and Qwen-VL-Chat \cite{bai2023qwenvl}. We use the official implementation to perform inference for each of these LMMs. Given that our approach is a Chain-of-Thought method, there is no necessity to make any adjustments to either the model architecture or the inference process. During inference, we set the temperature parameter to 0 in all experiments for optimal performance and stability of the model output, except for the experiment for CoT-SC \cite{CoT-sc}, where we set it to 0.7 according to its original implementation. 
In the following, we briefly review these LMMs.

\smallskip\textbf{LLaVA-1.5.} The model architecture of LLaVA-1.5 \cite{llava1.5}  replaces the linear projection with an MLP to map visual features into a shared embedding space with the LLM. LLaVA-1.5 \cite{llava1.5} used CLIP-ViT-L~\cite{clip} as the visual encoder at 336$\times$336 resolution, and the LLM Vicuna \cite{chiang2023vicuna} functioned as the language decoder. LLaVA-1.5~\cite{llava1.5} employed region-level VQA datasets (Visual Genome \cite{visualgenome}, RefCOCO \cite{refcoco_1, refcoco_2}) to enhance the model’s capability of localizing fine-grained visual instances.
In our work, we utilize both the LLaVA-1.5-7B~\cite{llava1.5} and the LLaVA-1.5-13B~\cite{llava1.5} models for evaluation to verify our TextCoT. We adopt the accuracy~\cite{multimodalocr} as the metric, where a response generated by the model is considered correct if it contains the string present in the ground truth.

\smallskip\textbf{SPHINX.} SPHINX \cite{lin2023sphinx} introduced a weight-mixing strategy to efficiently combine domain-specific knowledge and unfreeze its LLM weights during instruction tuning. SPHINX \cite{lin2023sphinx} has broader multimodal question-answering tasks including region-level understanding, caption grounding, and document layout detection. We adopt the same metric as LLaVA-1.5 \cite{llava1.5} in the experiments.

\smallskip\textbf{ShareGPT4V.} ShareGPT4V-7B \cite{sharegpt4v} model follows the design of LLaVA-1.5~\cite{llava1.5}. It incorporated the ShareGPT4V dataset into both the pre-training and SFT phases. With 7B parameters, ShareGPT4V-7B \cite{sharegpt4v} outperforms competitors with a remarkable performance across a majority of the multimodal benchmarks, despite these competitors using larger training datasets or more parameters. We adopt the same evaluation metric as LLaVA-1.5 \cite{llava1.5}.

\smallskip\textbf{Qwen-VL-Chat.} Qwen-VL-7B \cite{bai2023qwenvl} and Qwen-VL-Chat-7B \cite{bai2023qwenvl} are a series of highly performant and versatile vision-language foundation models based on the Qwen-7B \cite{qwen} large language model. The LLM basement is empowered with visual capacity by introducing a new visual receptor including a language-aligned visual encoder and a position-aware adapter. 
For this model, we opt for the same accuracy metric \cite{multimodalocr} mentioned above.

\smallskip\textbf{GPT-4V.} Unlike the previous models, the architecture and pretraining details of GPT-4V \cite{gpt4} are not made public. Some technical reports \cite{gpt4dawn} revealed the extraordinary performance of GPT-4V \cite{gpt4} including captioning, object localization, and counting. GPT-4V \cite{gpt4} demonstrates the capability to generate bounding box coordinates in textual format, without separate textualized box tokens.

\smallskip\textbf{Claude.} The architecture and pretraining details of Claude 3 \cite{claude} are not made public. Claude 3 \cite{claude} possesses the capability to generate bounding box coordinates in textual format. We use the Claude 3 Opus \cite{claude} model for our qualitative experiment, which is the most powerful version.

\begin{table*}[t]
\setlength{\tabcolsep}{1.65mm}
\centering
\small
  \caption{
  Qualitative comparison between two baseline LMMs and those augmented by different CoT methods.}
  \label{tab:CoTtable}
  
  \begin{tabular}{l|cc|ccc|ccc|c}
    \toprule
    \multirow{2}{*}{Method} & \multicolumn{2}{c}{Scene Text-Centric VQA} &  \multicolumn{3}{c}{Document-Oriented VQA} & \multicolumn{3}{c|}{KIE} & \multirow{2}{*}{Average} \\
     & TextVQA & STVQA  & DocVQA & InfoVQA & ChartQA & POIE & SROIE & FUNSD\\
    \midrule
    LLaVA-1.5-7B \cite{llava1.5} & 37.92 & 38.85 & 8.97 & 14.71 & 6.40 & \textbf{2.61} & 1.73 & 0.46 & 13.96\\
    LLaVA-1.5-7B + ZS-CoT \cite{zs-CoT} & 36.11 & 35.93 & 9.04 & 15.25 & 6.29 & 1.67 & 3.60 & 1.16 & 13.63\\
    LLaVA-1.5-7B + CoT-SC \cite{CoT-sc} & 36.27 & 36.82 & 8.60 & 15.01 & 6.18 & 1.79 & 2.67  & 1.62 & 13.62\\
    LLaVA-1.5-7B + DDCoT \cite{zheng2023ddCoT} & 37.77 & 38.71 & 8.77 & 14.06 & \textbf{6.85} & 2.36 & 1.95 & 0.46 & 13.87\\
    LLaVA-1.5-7B + CCoT \cite{cCoT} & 38.44 & 37.91 & 9.39 & 13.88 & 6.07 & 2.14 & 3.03 & 1.16 & 14.00\\
    LLaVA-1.5-7B + \textbf{TextCoT} & \textbf{46.08} & \textbf{40.69} & \textbf{11.35} & \textbf{15.43} & 6.74 & 2.20 & \textbf{6.20} & \textbf{3.02} & \textbf{16.46}\\
    \midrule
    ShareGPT4V \cite{sharegpt4v} & 43.10 & 43.75 & 11.56 & 15.19 & 8.65 & 10.97 & 6.84 & 1.86 & 17.74\\
    ShareGPT4V + ZS-CoT \cite{zs-CoT} & 41.48 & 42.20 & 11.56 & 15.55 & 9.33 & 9.84 & 7.20 & 1.39 & 17.32\\
    ShareGPT4V + CoT-SC \cite{CoT-sc} & 42.27   & 42.76 & 11.37 & 16.69 & 7.75 & 7.95 & 6.20 & 1.86 & 17.11\\
    ShareGPT4V + DDCoT \cite{zheng2023ddCoT} & 41.88 & 42.57 & 11.19 & 15.07 & 10.11 & 10.53 & 6.99 & 2.09 & 17.55\\
    ShareGPT4V + CCoT \cite{cCoT} & 42.34 & 43.89 & 11.42 & 14.65 & 9.44 & 10.37 & 7.85 & 1.62 & 17.70\\
    ShareGPT4V + \textbf{TextCoT} & \textbf{52.79} & \textbf{46.68} & \textbf{16.89} & \textbf{16.99} & \textbf{12.13} & \textbf{12.54} & \textbf{13.69} & \textbf{13.92} & \textbf{23.20}\\ 
    \bottomrule
  \end{tabular}
\end{table*}

\subsection{Datasets}
To illustrate the strong generalization of our TextCoT,
we choose several datasets for evaluation that cover a wide range of scenarios,
including Scene Text-Centric Visual Question Answering, Document-Oriented VQA, and Key Information Extraction datasets. In the following, we provide a brief introduction to these datasets.

\smallskip\textbf{Scene Text-Centric VQA.} 
TextVQA \cite{textvqa} and STVQA \cite{stvqa} are two of the most commonly used benchmark datasets in the field of Scene Text-Centric VQA. 
The TextVQA \cite{textvqa} benchmark dataset contains over 45,000 questions for 28,000 images. These images are selected from various categories in the OpenImages \cite{krasin2017openimages} dataset. STVQA \cite{stvqa} benchmark dataset comprises over 31,000 questions for 23,000 images collected from diverse public datasets.

\smallskip\textbf{Document-Oriented VQA.} 
DocVQA \cite{docvqa}, InfographicVQA \cite{mathew2022infographicvqa}, and ChartQA \cite{masry2022chartqa} are three widely-used benchmark datasets for Document-Oriented VQA tasks. The DocVQA \cite{docvqa} dataset consists of 12,767 document images of various types and content, alongside over 50,000 associated questions and answers. InfographicVQA \cite{mathew2022infographicvqa} dataset comprises a diverse collection of 5,485 infographic images, along with a cumulative total of 30,035 questions. ChartQA \cite{masry2022chartqa} dataset encompasses 9,608 manually crafted questions addressing 4,804 charts and 23,111 questions generated from human-written summaries of 17,141 charts.

\smallskip\textbf{Key Information Extraction (KIE).} We further use three prevalent datasets commonly employed in the field of Key Information Extraction: SROIE \cite{sroie}, FUNSD \cite{funsd}, and POIE \cite{poie}. The SROIE \cite{sroie} dataset consists of 1000 scanned receipt images specifically designed for OCR and key information extraction competitions. Participants are tasked with identifying essential details such as company names, issuance dates, addresses, and total expenditures from these receipts. The FUNSD \cite{funsd} dataset provides a collection of 199 authentic, fully annotated scanned forms, which may contain instances of noise. This dataset poses a unique challenge due to its real-world variability and potential for ambiguity. The POIE \cite{poie} dataset concentrates on Nutrition Facts labels of products in English, amassing a significant repository of 3,000 images containing a total of 111,155 instances of text. The primary objective here is to extract relevant information from these labels.

\subsection{Results}

We first compare our method with the baseline LMMs on the above question-answering datasets. Then, we compare our method with previous CoT methods.

\smallskip\textbf{Quantitative Comparison with Baseline LMMs.}
In Table \ref{tab:main_table}, we evaluate the performance of the five baseline LMMs as well as the performance after integrating our TextCoT.
The evaluated LMMs involve various model sizes, training data, and architectures.

Firstly, our TextCoT significantly enhances the performance of five advanced LMMs on almost all eight datasets. This result verifies our idea of progressively scrutinizing local details to furnish more accurate responses.
Secondly, comparing LLaVA-1.5-7B \cite{llava1.5} and LLaVA-1.5-13B \cite{llava1.5} with identical architectures but different model size, our TextCoT achieves an average accuracy improvement of 2.51\% and 3.72\%, respectively. It is noteworthy that the larger models benefit more from our TextCoT. A plausible explanation for this phenomenon is that larger models possess enhanced cognitive and reasoning capabilities, resulting in greater benefits. This conclusion has also been substantiated within the domain of LLMs~\cite{CoT}.

Thirdly, contrasting LLaVA-1.5-7B \cite{llava1.5} with ShareGPT4V \cite{sharegpt4v} that uses high-quality captioned data but the same model size and architecture, our TextCoT produces average accuracy improvements of 2.5\% and 5.46\%, respectively. 
The comparison indicates that the effectiveness of TextCoT increases with 
better text-image alignment in the model, associated with higher-quality training data. 
Lastly, when applying TextCoT to the Qwen-VL-Chat~\cite{bai2023qwenvl} with different architecture and training data from the above LMMs, performance improvements are observed on most datasets. 
However, a performance decrease is observed on DocVQA~\cite{docvqa} and ChartQA~\cite{masry2022chartqa} datasets. This can be attributed to the fact that the Qwen-VL-Chat~\cite{bai2023qwenvl} was trained on these datasets, causing it to become overly familiar with their questioning style and image characteristics. 
Consequently, when attempting to alter the questioning style through the application of TextCoT, a performance decline is observed.

\begin{figure}[t]
  \centering
  \includegraphics[width=\linewidth]{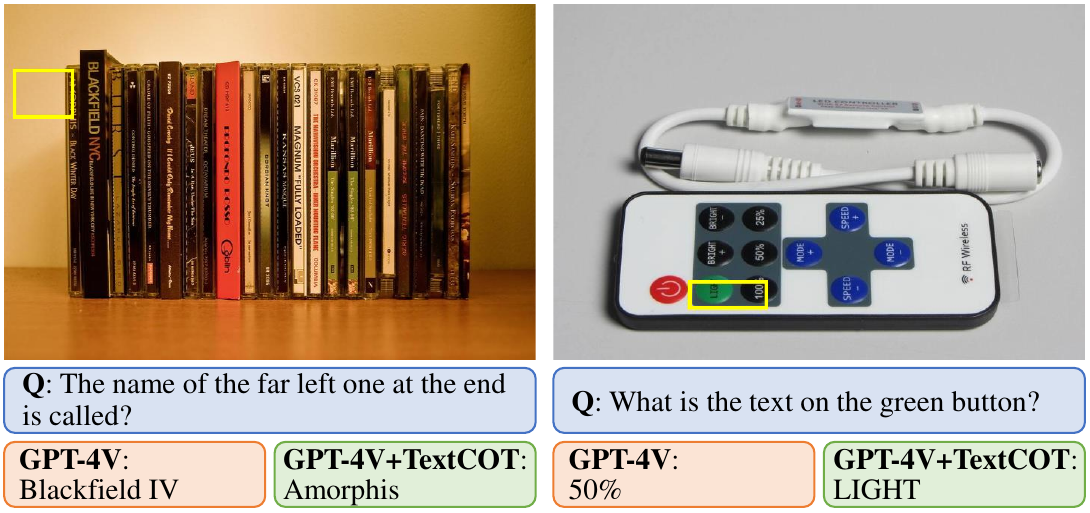}
  
  \caption{
        Comparison between the responses of GPT-4V~\cite{gpt4} and those augmented by our TextCoT on several cases. The estimated answer regions \(A_g\) in the second stage are highlighted in the image using yellow bounding boxes.
  }
  \label{fig:gpt4_case}
\end{figure}

\begin{figure}[t]
  \centering
  \includegraphics[width=\linewidth]{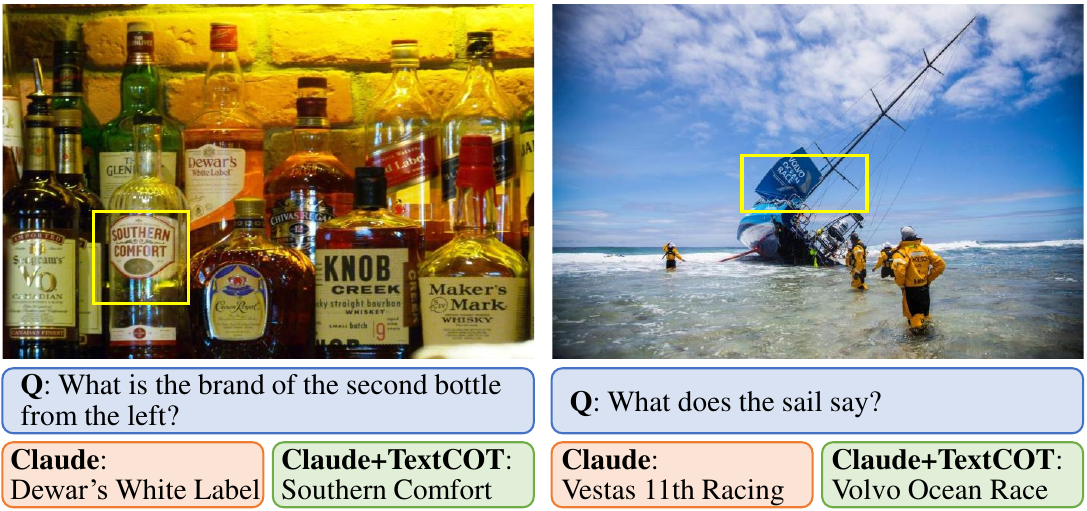}
  
  \caption{
          Comparison between the responses of Claude~\cite{claude} and those augmented by our TextCoT on several cases. The estimated answer regions \(A_g\) in the second stage are highlighted in the image using yellow bounding boxes.
  }
  \label{fig:claude_case}
  
\end{figure}

\smallskip\textbf{Qualitative Comparison with Baseline LMMs.}
We further conduct a series of qualitative comparisons. Initially, as illustrated in Figure \ref{fig:scenetext_case_1}, Figure \ref{fig:scenetext_case_2}, and Figure \ref{fig:document_case}, we present the responses of two baseline LMMs~\cite{llava1.5,sharegpt4v} as well as the enhanced responses after the integration of our TextCoT. As we can see, our method succeeds in approximating the answer regions within the images and rectifying the inaccurate responses of the baseline LMMs in both scene text and document scenarios. In addition, we also conduct experiments on the advanced GPT-4V \cite{gpt4} and Claude 3 Opus \cite{claude} models. As shown in Figure~\ref{fig:gpt4_case} and Figure~\ref{fig:claude_case}, our TextCoT enhances the accuracy of the responses for the two models. 
While the bounding box positions given by LMMs are inaccurate, our TextCoT method remains unaffected by such discrepancies.

\smallskip\textbf{Comparison with Previous CoT Methods.} We also conduct a performance comparison of existing CoT methods based on LLAVA-1.5-7B \cite{llava1.5} and ShareGPT4V \cite{sharegpt4v}. Comparison methods include ZS-CoT~\cite{zs-CoT} and CoT-SC~\cite{CoT-sc} for LLMs, as well as DDCoT~\cite{zheng2023ddCoT} and CCoT~\cite{cCoT} for LMMs.
ZS-CoT~\cite{zs-CoT}, DDCoT~\cite{zheng2023ddCoT} and CCoT~\cite{cCoT} consist of two stages, where ZS-CoT~\cite{zs-CoT} utilizes a ``Let's think step-by-step'' approach. For CoT-SC~\cite{CoT-sc}, we sample 5 Chain-of-Thought reasoning paths. 

In Table \ref{tab:CoTtable}, the results demonstrate that our approach significantly outperforms these methods. 
In text-rich image scenarios, these methods fail to improve performance. The reason is that they did not solve the need for local and fine-grained visual inputs. In contrast, our approach effectively leverages the captioning and localization capabilities of Large Multimodal Models (LMMs) to extract both global and local information for accurate answering.   

\begin{table*}[t]
\setlength{\tabcolsep}{1.65mm}
\centering
\small
  \caption{
  Ablation studies about the reasoning process in our TextCoT.
  The term ``ground'' denotes the generation of the answer region \(A_g\) without executing a cropping operation.
  The experiments are conducted on LLaVA-1.5-7B~\cite{llava1.5}.}
  \label{tab:ablation}
  
  \begin{tabular}{c|ccc|cc|ccc|ccc|c}
    \toprule
     \multicolumn{4}{c|}{Method} & \multicolumn{2}{c}{Scene Text-Centric VQA} &  \multicolumn{3}{c}{Document-Oriented VQA} & \multicolumn{3}{c|}{KIE}  & \multirow{2}{*}{Average} \\
   \multicolumn{1}{c}{} & Ground & Crop & Caption & TextVQA & STVQA & DocVQA & InfoVQA & ChartQA & POIE & SROIE & FUNSD\\
    \midrule
     (a) & & & & 37.92 & 38.85 & 8.97 & 14.71 & 6.40 & 2.61 & 1.73 & 0.46 & 13.96\\
     (b) & \checkmark & & & 39.39 & 37.95 & 9.46 & \textbf{15.43} & 6.97 & \textbf{2.64} & 4.25 & 0.23 & 14.54 \\
     (c) & \checkmark &  \checkmark & &  45.52 & 40.64& 11.10 & 14.89 & \textbf{7.08} & 2.39 & 5.91 & 2.32 & 16.23\\
     (d) & \checkmark  & \checkmark  & \checkmark  &  \textbf{46.08} & \textbf{40.69}& \textbf{11.35} & \textbf{15.43} & 6.74 & 2.20 & \textbf{6.20} & \textbf{3.02} & \textbf{16.46}\\
    \bottomrule
  \end{tabular}
\end{table*}

\begin{table*}[t]
\setlength{\tabcolsep}{1.65mm}
\centering
\small
  \caption{
  Ablation studies about the cropping strategies in our TextCoT. The setting used in our final method is underlined. The experiments are conducted on LLaVA-1.5-7B~\cite{llava1.5}.
  }
  \label{tab:hyperparameters}
  
  \begin{tabular}{c|c|cc|ccc|ccc|c}
    \toprule
    \multicolumn{2}{c|}{\multirow{2}{*}{Cropping}} &  \multicolumn{2}{c}{Scene Text-Centric VQA} &  \multicolumn{3}{c}{Document-Oriented VQA} & \multicolumn{3}{c|}{KIE} & \multirow{2}{*}{Average} \\
    \multicolumn{2}{c|}{}& TextVQA & STVQA  & DocVQA & InfoVQA & ChartQA & POIE & SROIE & FUNSD\\
    \midrule
    (a) & Baseline & 37.92 & 38.85 & 8.97 & 14.71 & 6.40 & \textbf{2.61} & 1.73 & 0.46 & 13.96\\
    (b) & Maintain rectangular & 29.92 & 24.75 & 7.36 & 13.58 & 4.04 & 0.53 & 1.51 & 0.70 & 10.30\\
    (c) & Expand to square & 45.52 & 40.59 & \textbf{11.46} & 14.89 & 6.29 & 2.23 & 3.96 & \textbf{3.94} & 16.11\\
    (d) & \underline{Expand to 1.5x square} & \textbf{46.08} & \textbf{40.69} & 11.35 & \textbf{15.43} & \textbf{6.74} & 2.20 & \textbf{6.20} & 3.02 & \textbf{16.46}\\
    (e) & Original image & 38.96 & 38.90 & 9.27 & 14.77 & 6.07 & 2.42 & 2.09 & 0.70 & 14.15\\
    
    \bottomrule
  \end{tabular}
\end{table*}

\subsection{Ablation Study}

To validate the effectiveness of our proposed TextCoT 
across three stages, 
we conduct in-depth ablation experiments as detailed in Table \ref{tab:ablation} and \ref{tab:hyperparameters}. All ablation experiments are conducted based on the classic LLaVA-1.5-7B~\cite{llava1.5} and the performance is assessed across eight question-answering datasets. We discuss the results next.

\smallskip\textbf{Impact of Coarse Localization.} 
We first evaluate a two-stage variant (Table~\ref{tab:ablation} (b)) of our TextCoT, where the first stage predicts a grounding answer \(A_g\), and the second stage directly feeds \(A_g\) into the LMM. Compared with the one-stage baseline method that takes image \(I_g\) and question \(Q\) as input (Table~\ref{tab:ablation} (a)), this variant performs better. The results highlight the significance of answer area cues. However, this variant has not fully leveraged answer area prompts, and the model still lacks local fine-grained visual inputs.

\smallskip\textbf{Impact of Cropping Process.} 
Based on this two-stage variant (Table~\ref{tab:ablation} (b)), 
we further introduce a cropping operation for images, allowing the LMM to obtain detailed local information.
As shown in Table \ref{tab:ablation}, the method incorporating a cropping operation (Table~\ref{tab:ablation} (c)) exhibits superior performance.
The significant improvement in text-related VQA tasks indicates that magnifying the answer regions significantly enhances the understanding of local details.

\smallskip\textbf{Impact of Image Overview.} 
Since the model loses global contextual information after performing the cropping operation, it can only extract local information from cropped image \(I_l\). To address this issue, we incorporate captions of the global image \(I_g\), namely our Image Overview stage. 
As shown in Table \ref{tab:ablation} (d), the supplementation of global information further enhances the performance.
This approach provides global information through textual descriptions while utilizing local images for detailed local information. This constitutes our final TextCoT.

\smallskip\textbf{Impact of Specific Cropping Method.} 
We further conduct a series of ablation studies on our cropping strategy. As shown in Table \ref{tab:hyperparameters}, we experiment with 
strict cropping according to the grounding area \(A_g\) (Table \ref{tab:hyperparameters} (b)), expanding the bbox to a square by aligning the longer side of the bbox (Table \ref{tab:hyperparameters} (c)), expanding the bbox to a square with side length equal to 1.5 times the longer side of the bbox (Table \ref{tab:hyperparameters} (d)), and not cropping at all (Table \ref{tab:hyperparameters} (e)).

The experimental results demonstrate that our TextCoT configuration (Table \ref{tab:hyperparameters} (d)) exhibits superior performance. This can be attributed to the fact that, in comparison to other cropping settings, our configuration retains sufficient fine-grained visual information while also avoiding the omission of answer regions due to insufficient cropping. This result is also consistent with our idea.

\section{Limitations}
We further discuss the limitations of our TextCoT.
Firstly, although our method demonstrates a tolerance for errors in the model's grounding capabilities, it is not compatible with certain LMMs that lack detection capabilities. Secondly, in some complex table images, answers often span across multiple distinct regions, which imposes a higher demand on the model's grounding capability. Exploring how to 
precisely extract local and fine-grained visual features for such models presents a meaningful research topic.
Thirdly, our evaluation is presently confined to the text domain, thus limiting the generalizability of our findings to other modalities or domains. While our current research focuses on text-related tasks, future endeavors will aim to develop CoT methods that can enhance performance across diverse and general scenarios beyond the text domain.

\section{Conclusion}

In this paper, we introduce TextCoT, a novel Chain-of-Thought framework tailored for the enhancement of the text-rich image understanding of LMMs. Our method enhances question-answering accuracy for text-rich images by leveraging the captioning and grounding capabilities of LMMs. This allows for the extraction of both global and local visual information. 
TextCoT offers seamless integration with existing LMM architectures, requiring no additional training and enabling immediate plug-and-play functionality.
Extensive experimentation across diverse text-rich image question-answering benchmarks based on several advanced LMMs has consistently demonstrated the efficacy and robust generalization capability of our TextCoT. 
Our work makes a significant stride toward unlocking the full potential of LMMs in comprehending text-rich visual data. In the future, we will focus on developing a method for LMMs even without grounding ability and enhance their ability to understand more comprehensive scenes.

{
    \small
    \bibliographystyle{ieeenat_fullname}
    \bibliography{main}
}

\end{document}